\pdfoutput=1

\documentclass[11pt]{article}

\usepackage{acl}

\usepackage{times}
\usepackage{latexsym}
\usepackage{graphicx}
\usepackage{CJKutf8}
\usepackage{arydshln}
\newcommand{\multirows}[1]{\begin{tabular}{@{}c@{}}#1\end{tabular}}

\usepackage[T1]{fontenc}

\usepackage[utf8]{inputenc}

\usepackage{microtype}

\title{Scenario-based Multi-product Advertising Copywriting Generation for E-Commerce}

\author {
    Xueying Zhang\textsuperscript{\rm 1},
    Kai Shen\textsuperscript{\rm 2}, 
    Chi Zhang\textsuperscript{\rm 1}, 
    Xiaochuan Fan\textsuperscript{\rm 1}, 
    Yun Xiao\textsuperscript{\rm 1}, 
    Zhen He\textsuperscript{\rm 2}, 
    Bo Long\textsuperscript{\rm 2},
    Lingfei Wu\textsuperscript{\rm 1$\dagger$} \\
    \textsuperscript{\rm 1} JD.COM Silicon Valley Research Center\\
    \textsuperscript{\rm 2} JD.COM\\
    lingfei.wu@jd.com
}

\begin{document}
\maketitle
\begin{abstract}

In this paper, we proposed an automatic Scenario-based Multi-product Advertising Copywriting Generation system (SMPACG)  for E-Commerce, which has been deployed on a leading Chinese e-commerce platform. The proposed SMPACG consists of two main components: 1) an automatic multi-product combination selection module, which itself is consisted of a topic prediction model, a pattern and attribute-based selection model and an arbitrator model; and 2) an automatic multi-product advertising copywriting generation module, which combines our proposed domain-specific pretrained language model and knowledge-based data enhancement model. The SMPACG is the first system that realizes automatic scenario-based multi-product advertising contents generation, which achieves significant improvements over other state-of-the-art methods. The SMPACG has been not only developed for directly serving for our e-commerce recommendation system, but also used as a real-time writing assistant tool for merchants.

\end{abstract}

\section{Introduction}

The Internet and mobile Internet develop rapidly over the last decade, online/mobile shopping has become a mainstay way for many people around the world. Advertising copywriting plays a significant role in e-commerce recommendation platforms, especially for these e-commerce technology giants like Taobao.COM, JD.COM, etc. In these e-commerce platforms, well-written advertising copywriting can be beneficial for attracting customers and further increasing sales. However, producing advertisements by human copywriters face a few of severe limitations: 1) the efficiency of human copywriters is not able to match the growth rate of products; 2) manpower cost becomes huge when number of products exponentially increases; 3) special training is required for different scenarios. Due to these reasons, automatic advertising copywriting generation has become an important and essential task in e-commerce.

\begin{figure}[t!]
\centering
\includegraphics[width=1\linewidth]{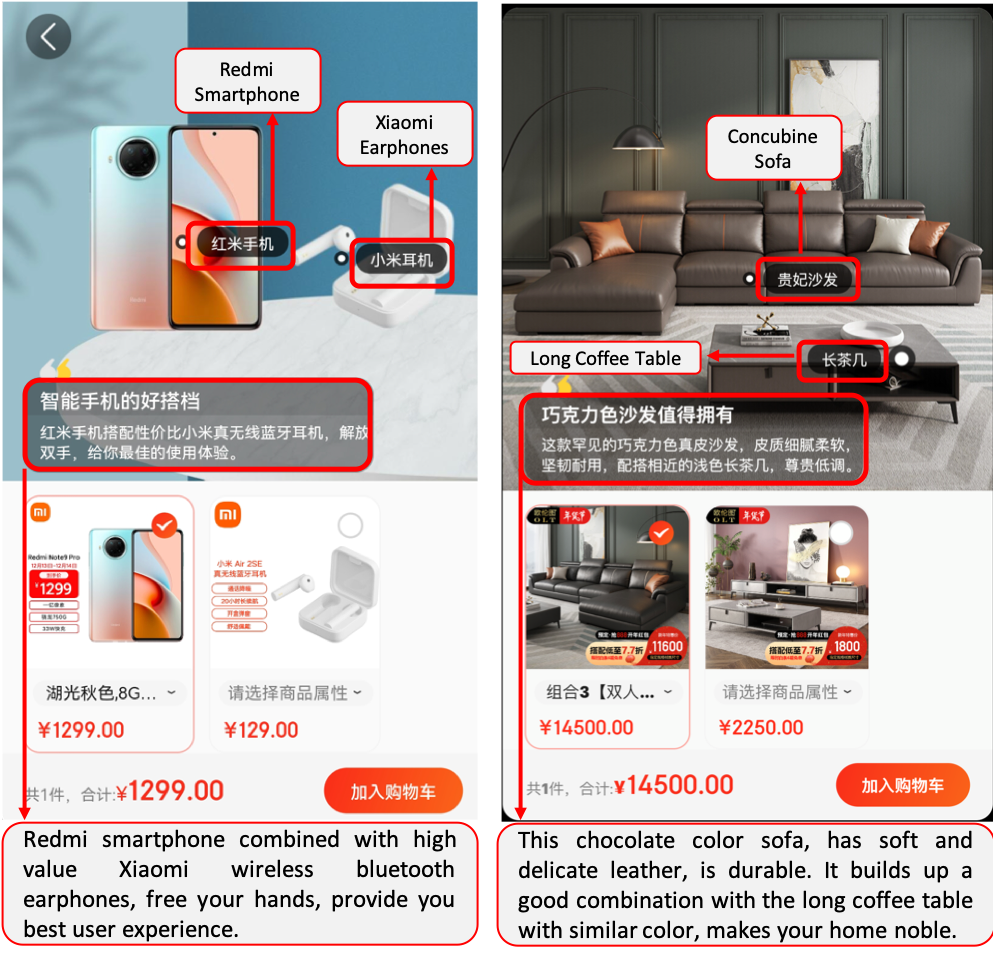}
\caption{Examples of multi-product advertising on our e-commerce platform}
\label{fig1}
\end{figure}

Recent years there has seen a surge of interests in automatic advertising copywriting generation \citep{ZhangZHR19,chen2019towards,zhang2021automatic,guo2021intelligent}, which however, most of them focus on single product description generation. Single product advertising copywriting typically focuses on only describing the characteristics of the product itself in order to help customers to better understand the product and facilitate potential sales. 
Single product advertising is the main sales mode in most of e-commerce platforms nowadays.
However, as the shopping needs and habits of customers become more diverse, displaying and selling multiple products as a combination has become a trending feature. In the multi-product mode, customers avoid troubles in finding essential related products. More importantly, e-commerce platforms gain significant more sales by encouraging customers to view and buy more related products.

\begin{figure*}[t!]
\centering
\includegraphics[width=1\linewidth]{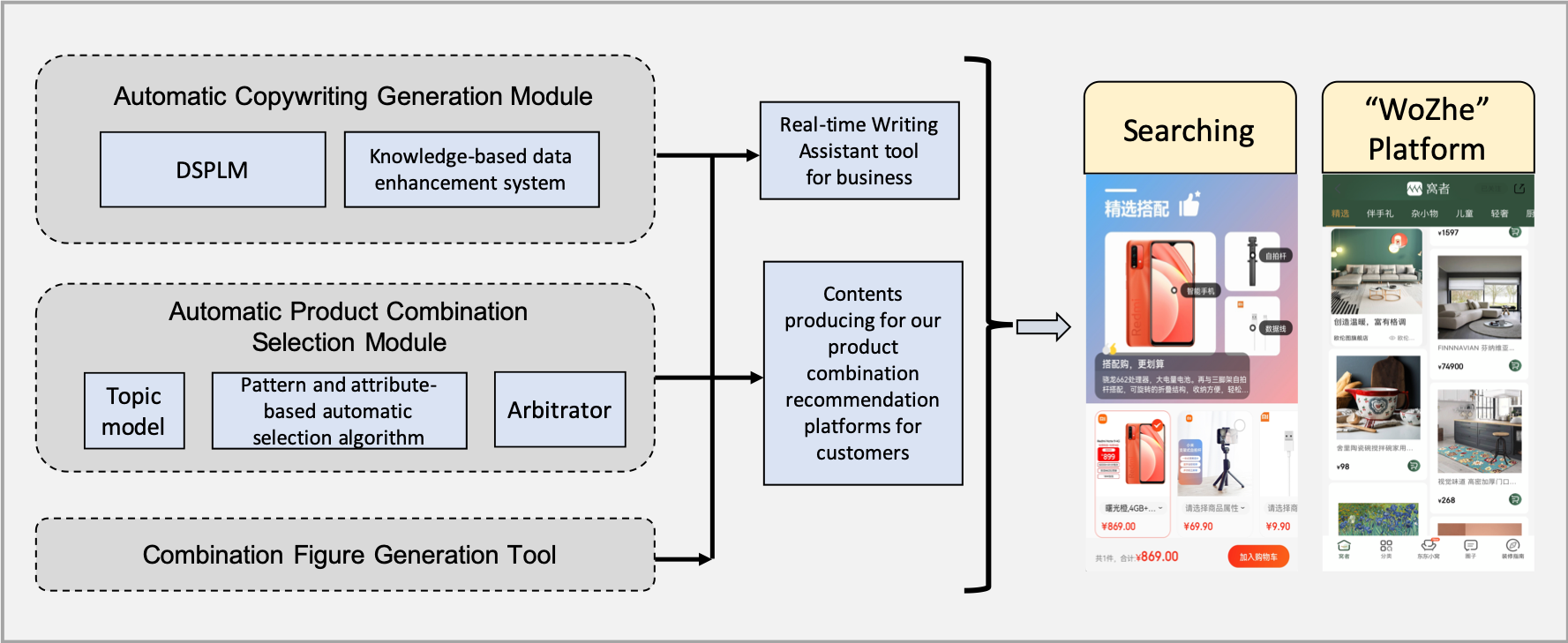}
\caption{Overview of SMPACG system}
\label{fig2}
\end{figure*}

There are few existing research works in the multi-product advertising copywriting generation. A recent work on multi-product advertising generation \cite{chan-etal-2020-selection} aimed to help customers make better selection among multiple products with similar characteristics, which however still generated one copywriting for one product and then simply assemble them together. In contrast, our proposed SMPACG is fundamentally different from theirs.  Our scenario-based multi-product advertising copywriting is developed to find highly related products (rather than similar products) under the certain scenario, and generates only one copywriting for multi-product to encourage customers to buy products within the combinations all together.
As shown in Figure 1, we encourage customers to buy the smartphone and earphones (or sofa and coffee table) together by highlighting how these two products work well together and why the customer should buy them together in our advertising copywriting.

To this end, the scenario-based multi-product combination advertising requires two essential parts: 1) an automatic product combination selection module, 2) an automatic multi-product advertising copywriting generation module. 
Product combination selection by human labors suffers similar issues as producing advertisements by human copywriters, and thus an automatic product combination selection algorithm is indispensable.
In contrast with the automatic copywriting generation system \cite{zhang2021automatic} in earlier work, which serves for the E-commerce platforms and applies on single product only, in our scenario, the generated descriptions need to be able to describe not only the information of each single product in the combination, but also the overall feature of the combination under the certain scenario. 
In addition, unlike the single product copywriting generation, multi-product copywriting generation requires an additional product combination selection module to select highly related yet not similar products. 

To overcome these challenges, we propose our automatic Scenario-based Multi-Product Advertising Copywriting Generation (SMPACG) system for e-commerce. The SMPACG is the first system that realizes automatic scenario-based multi-product advertising contents generation, which achieves significant improvements over other state-of-the-art methods. The SMPACG has been not only developed for directly serving for our e-commerce recommendation system, but also used as a real-time writing assistant tool for merchants.

\section{Related Work}

Natural language generation focuses on constructing computer systems that can generate understandable texts in human languages from some underlying textual or non-linguistic representation of information \cite{reiter_dale_2000}. 
In recent years, NLG techniques have been developing rapidly and applied to a wide range of applications such as abstract text summarization \cite{rush2015neural,michele2000headline,kevin2000statistics}, 
dialogue generation \cite{ijcai2018-614,zhang2020dialogpt},
machine translation \cite{bahdanau2016neural} and
creative generation \cite{Yao_Peng_Weischedel_Knight_Zhao_Yan_2019,li-etal-2018-generating-classical}.

At early stage of text generation research, since data and computational resources are limited, models relied heavily on feature engineering \cite{laffertyCrf,och-etal-2004-smorgasbord}.
As the technology of deep learning develops, features are learned automatically and defined implicitly during model training. 
The natural language generation problem has been framed as a sequence-to-sequence task and various studies have been proposed to enhance the model architecture,  
such as convolutional neural networks (CNN) \cite{gehring2017convolutional}, recurrent neural networks (RNN) \cite{sutskever2014sequence}, graph neural networks (GNN) \cite{xu2018graph2seq,li2020graph},
and Transformer-based architectures \cite{bahdanau2016neural,vaswani2017attention}. 

Text generation techniques have been widely used for automatic product description generation tasks. Early work focuses on applying template-based generation methods \cite{wang2017statistical}, while the template coverage and diversity are limited.
\citet{ZhangZHR19} proposed a pointer-based generation model with a dual encoder to generate product description.
\citet{chen2019towards} proposed a external knowledge based transformer-based generation model, which utilized relevant information from customer reviews extracted by an adaptive posterior distillation module.
\citet{zhang2021automatic} presented the proposed automatic product copywriting generation system for single-product, which has been employed for the leading e-commerce platform. 
\citet{guo2021intelligent} developed and deployed the intelligent online selling point extraction system to serve the recommendation system on the leading Chinese e-commerce platform. 
\citet{chan-etal-2020-selection} designed SMGNet to generate the advertising post for multi-product scenarios which consists of similar products for ease of choice for customers.

\section{Approach}



In this section, as shown in Figure 2, we first illustrate the overall architecture of the SMPACG system. Our system consists of an automatic multi-product advertising copywriting generation module that itself also serves as a real-time writing assistant tool for merchants, the automatic multi-product combination selection module and combination figure generation tool for directly serving for our e-commerce recommendation system, including various applications such as the searching channel and the ``WoZhe'' platform (mainly for home appliance).
We will discuss SMPACG system in details in the following sections.

\subsection{Data Collection}
We built a real-world dataset from our in-house platform which includes three parts: 
1) product combination data, 
2) product combination advertising copywriting data, 
and 3) product information data.
The product combinations and corresponding advertising copywriting are selected and written by professional copywriters with good knowledge of marketing. 
Within each product combination, highly related products under the certain scenario are selected, and the corresponding advertising copywriting includes both advertising content and title. 
For product data, we collect multiple
types of data from our in-house database, including
product titles, attributes and product detail images.

\subsection{Automatic Product Combination Selection Module}

\subsubsection{Topic Channel Generation}
We build a system to cluster our products into different topic channels based on their characteristics. The details of this topic channel system are not included in this work, it is still under development and will be discussed in more details in our further work. The first version of this system is adopted in our work to further build our automatic product combination selection algorithm.

\subsubsection{Pattern and Attribute-based Automatic Selection}
Given that we have products already labeled to certain topics, the simple approach is to randomly select products within the topics to build the combination. We use this approach as our baseline. Random selected combinations within the same topic may suffer some issues, for instance, products with same functions are selected as combination, or products with no obvious relationship are selected. In order to overcome these obstacles, we propose the pattern and attribute-based automatic selection algorithm. We extract attribute patterns from our product combination dataset, and map these patterns to each of our topic channels, then the product combinations are randomly selected within the mapped attribute patterns. The attributes we utilize to build the patterns include product detailed category and the product word predicted from the product title with the product word prediction model. We utilize the product word prediction model developed in our other projects here.

\begin{figure}[t!]
\centering
\includegraphics[width=1\linewidth]{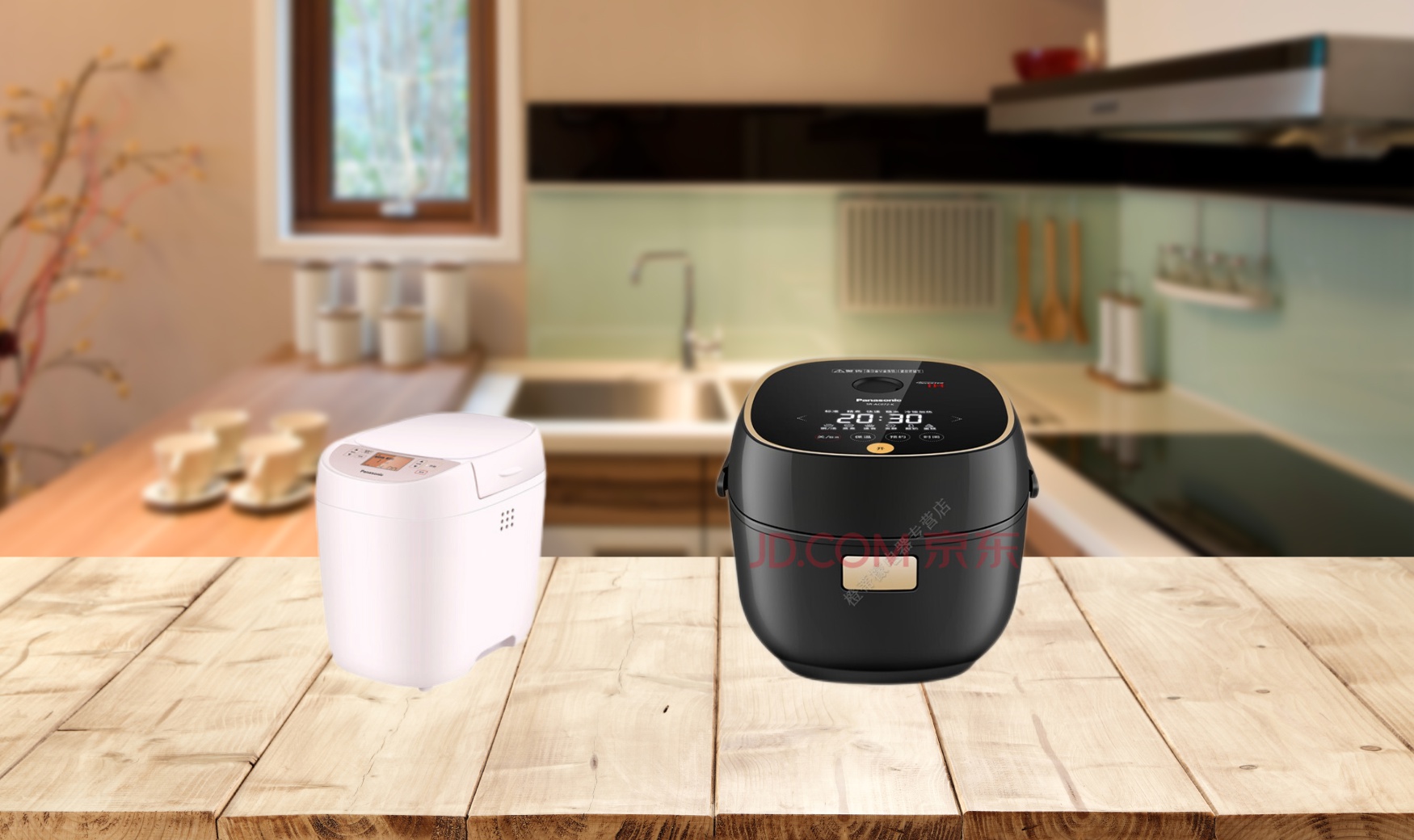}
\caption{Examples of figures generated by combination figure generation tool of SMPACG system}
\label{fig5}
\end{figure}

\subsubsection{Arbitrator}

Once we obtain the generated combinations, we need an arbitrator for further evaluation and selection of our selected combinations. We train two BERT-based classifiers\cite{devlin-etal-2019-bert} as our arbitrators, the difference between these two arbitrators come from selection of negative samples. Both arbitrators are used for evaluation of our product combination selection algorithms, and the strict one acts as the filter for generating final results. 

By introducing the pattern and attribute-based automatic selection algorithm and the arbitrator, our automatic product combination selection module gain the ability of automatically generation of combinations with highly related products.
As shown in Figure 3, after obtaining the combination, combination figure generation tool is applied for producing a combination figure. 
The combination selection module and figure generation tool serve as a significant part of our SMPACG system.


\subsection{The DSPLM Model}


\subsubsection{Prefix Language Model}
The prefix LM \cite{dong2019unified} is a left-to-right Language Model that decodes b on condition of a prefixed sequence a, which is encoded by the same model parameters with a bi-directional mask. 
As shown in Figure 4, the input tokens can attend to each other bidirectionally, while the output tokens can only attend to tokens on the left. 
To realize this, a corrupted text reconstruction objective is usually applied
over input tokens, and an auto regressive language modeling objective is applied over output tokens, which encourages the prefix LM to better learn representations of the input.

The flexible model structure of prefix LM makes it a good choice for our application, where the power of language is utilized to realize our special training goals. Language can specify different parts of inputs and outputs as a sequence of symbols in a flexible way. Multiple kinds of information are embedded into the inputs represented by languages with specific tricks in building up the dataset, and the language model will learn the tasks implicitly. In this way, one model with the same structure can be reused for different tasks by only changing the input data which contains information of inputs, outputs and tasks all together.

\subsubsection{Backbone Network}

We utilize a 12-layer Transformer as our backbone network, the input vectors are encoded to contextual representations through the Transformer blocks. Each layer of Transformer block contains multiple self-attention heads, which takes the output vectors of the previous layer as inputs.

In our model, different mask matrices are fed into the model depending on the inputs. These mask matrices help control what context a token can attend to when calculating its contextual representation. As shown in Figure 4, within the input segment, the mask matrix is bidirectional, where the token can attend to both information before and behind. While for the output segment, the mask matrix is diagonal, where the current token can only attend to the tokens before.

\begin{figure}[t!]
\centering
\includegraphics[width=1\linewidth]{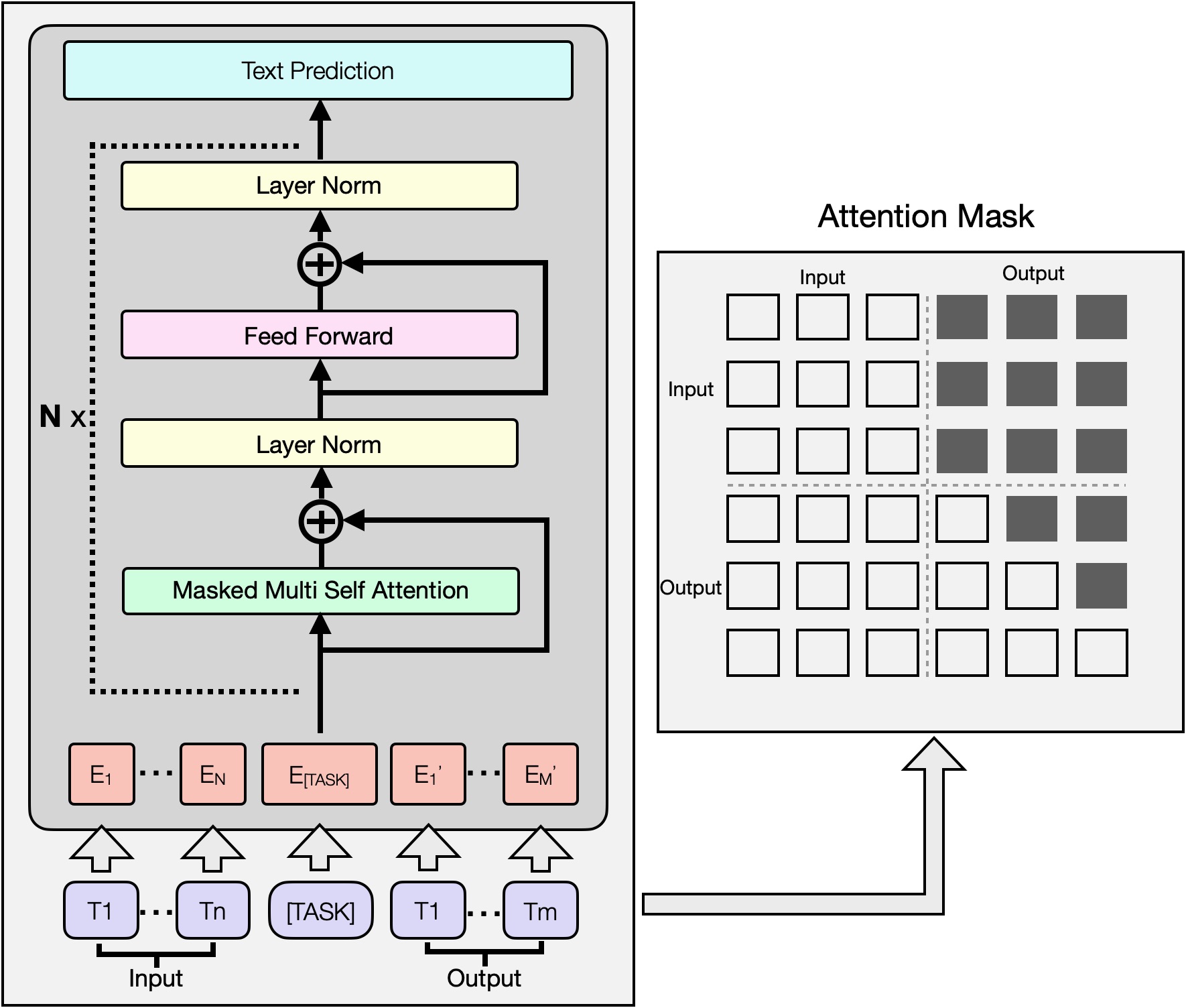}
\caption{Model structure and attention mask}
\label{fig4}
\end{figure}

\subsubsection{Domain-Specific Pretraining for Prefix LM}

Due to the limitation of training data, our model sometimes suffer the issues of generating descriptions with bad fluency or inaccurate information. In this case, our model never see any similar of related information before, which makes it hard for the model to generate accurate descriptions. 
Thus, we introduce pretraining into our prefix LM to solve the problem. Usually, when talking about pretraining, we focus on those open-domain models obtained with large amounts of data and compute resources. But this kind of pretraining models consume large training efforts, and are not applicable for serving online in production.
As shown in recent work \cite{zhang2021dsgpt}, domain-specific pre-training is a good idea to solve the problem, where we can also get benefits from pre-training with small amounts of data and small models. 
In this paper, we propose the DSPLM (Domain-Specific Pretrained prefix Language Model). We collect domain specific data from our in-house dataset, parts of our single product descriptions are utilized here. Although the task of single-product and multi-product copywriting are different, they are within the same domain: e-commerce copywriting. The domain-specific pretraining helps the model learn the e-commerce copywriting styles and obtain knowledge of a large range of products.

\subsection{Knowledge-based data enhancement}

At early stage of experiments, we observe that although the machine evaluation metrics are improved significantly by introducing new model architectures, human evaluation scores are low. The reason is that the earlier collected copywritings written by human experts suffer several issues. Through human screening, we classify the data issues to three main categories: 1) containing forbidden patterns, 2) limit coverage of product information, 3) too simple. To overcome these issues, we introduce our knowledge-based data enhancement model to ensure the quality of training data.  
Our data enhancement model consists of the following components: 
1) Avoided pattern filtering and cleaning: filtering or altering copywritings with certain patterns. 
2) Product word based checking model: ensuring the advertising copywriting cover information of all products in the combination. 
3) Creative information checking model: avoiding the advertising copywriting from simply listing all products.

Regarding the forbidden patterns, we divide them to two groups: alterable and non-alterable. We filter out the data with non-alterable avoided patterns and modify those with alterable patterns to build new dataset. 

We introduce the product word prediction model mentioned in Section 3.2.2 to our system for solving issues of limit coverage. The product word prediction model is a Bert-based model, with inputs of product title and other attributes, it can predict a group of matched product words with difference confidence scores. Utilizing these predicted product words, we build a rule-based algorithm to classify whether a advertising copywriting covers enough information of all products within the combination. Copywritings which don't meet the criterion are filtered out from the dataset. 

However, some copywritings cover information from all the products in the combinations by simply listing their titles or attributes, which we regard as too simple for using as an advertising copywriting. We build a rule-based creative information checking model to filter out these kinds of data. Our rule-base algorithm classify the advertising copywriting as too simple when no enough additional information beyond the product titles or attributes is provided in the copywriting.


\begin{table*}
\centering
\begin{tabular}{lllllllll}
\hline
Model & SacreBLUE & ROUGE-1 & ROUGE-2 & ROUGE-L & BLEU-1 & BLEU-4 & Meteor \\
\hline
T5-base    & 5.07     & 10.36    & 3.02  & 9.13   & 19.18   & 4.89    & 11.10  \\
\hline
Prefix LM    &16.15 & 28.92     & 12.96    & 24.86   & 35.86   & 16.53   & 20.49  \\
\hline
DSPLM    & \textbf{16.97}     & \textbf{29.36}    & \textbf{13.76}   & \textbf{25.24}   & \textbf{37.23}   & \textbf{17.34}    & \textbf{21.47} \\ 
\hline
\end{tabular}
\caption{Comparison of metrics of different generation methods}
\end{table*}

\begin{CJK*}{UTF8}{gbsn}
\begin{table*}
\centering
\scriptsize
\begin{tabular}{cccc}
\hline
Product Combination & Original Description & Issues & Generated Description \\
\hline
真皮沙发+玻璃茶几 & 
\multirows{
在繁忙的工作中，有这么一小片\\
恬静的空间，带走城市的喧嚣，\\
留下一地的静谧。}     & Limit coverage of products  & 
\multirows{
城市工作的人们，面临城市的喧嚣，\\
激烈的工作环境，让自己不放松。\\
选择简约的皮艺沙发，搭配简约玻璃 \\
茶几组合，让家具有时尚的魅力。}  \\
\hdashline
leather sofa + coffee table & 
\multirows{
In your busy life, there is a place that\\ 
takes away the hustle and bustle of the \\
city, and leaves you a space of peace.}     
& & \multirows{
People working in the city face the hustle \\
and bustle, intense working environment makes\\
yourself not relaxed. Choose a leather sofa \\
and a glass coffee table with concise design, \\
so that the furniture has a fashionable charm. }  \\
\hline
净水器+空气炸锅   & 再加99元享超大容量空气炸锅。 & 
 \multirows{ Too simple /\\ Contains forbidden pattern } & \multirows{易开得矽藻陶瓷净水器，滤除水中杂质，\\ 饮水安全，搭配空气炸锅，\\ 健康好物，美好生活！} \\
\hdashline
water filter + air fryer   & 
\multirows{
Add 99 dollar to enjoy an air \\
fryer with large capacity。} &  & \multirows{
The diatom ceramic water purifier, \\
filters out impurities in water，\\ 
lets you drink safely，together with \\
the air fryer，brings you \\ 
a healthy and happy life ！} \\
\hline
\end{tabular}
\caption{Effects on knowledge-based data enhancement model}
\end{table*}
\end{CJK*}


\section{Results}

In this section, we first compare our approach with several baselines and then provide case studies.

\subsection{Comparisons of Text Generation Models}

We evaluate our proposed DSPLM discussed in Section 3.3.3 on our in-house multi-product advertising copywriting dataset. For comparison, we train a T5-base model on the same dataset as baseline, and a Prefix LM to demonstrate the effectiveness of our proposed model. 
The T5-base model is randomly initialized due to the lack of pretrained models in Chinese, while the Prefix LM is initialized with Chinese Bert \cite{turc2019wellread}.
Table 1 shows evaluation metrics on these three different methods. The DSPLM
exceeds performance of Prefix LM and the T5-base model. Our DSPLM and PLM have 12 layers of transformer blocks each, while the T5-base model consists of a 6-layer encoder and 6-layers decoder, the total number of parameters are the same for these three models. The results are generated with beam search algorithm with beam size of 3.

\subsection{Effects of Knowledge-based Data Enhancement System}

In section 3.4, we discussed the knowledge-based data enhancement system proposed to ensure data quality.
To better illustrate the effects of this system, we compare some
examples of generated copywriting with the original data. 
As show in Table 2, the generated descriptions no long suffer the issues of the original data. 
For the first combination of sofa and coffee table, the original description fails to include key information of products, while the generated description successfully mentions both sofa and coffee table. 
For the second combination of water filter and air fryer, the original description not only contains forbidden pattern of promotion but also is too simple. The generated description doesn't contain any promotion words, in addition, it includes creative information from both the air-fryer and water filter instead of simply listing them.

By introducing the knowledge-based data enhancement system into our advertising copywriting generation module, after several rounds of optimization, the human screening approval rate increases from the original 20\% to 88\%.



\begin{table}
\centering
\begin{tabular}{llll}
\hline
Arbitrator & random & cid-based & pattern \\
\hline
Strict    & 6.56     & 6.67    & \textbf{16.41}  \\
\hline
Normal   & 5.09 & 10.29     & \textbf{20.16}   \\
\hline
\end{tabular}
\caption{Comparison of accuracy of different combination selection algorithms}
\end{table}

\subsection{Comparisons of automatic combination selection algorithms}

For evaluation of the proposed pattern and attribute-based automatic selection algorithm discussed in Section 3.2.2, we compare results of our method with two baselines. The first baseline is random selection within the topic channel, and the second one "cid-based" is random selection within detailed category
within the topic channel. Table 3 shows comparison of accuracy score of different combination selection algorithms, where the scores are predicted with the two arbitrators we discussed in Section 3.2.3. We can observe significant accuracy improvement of our proposed method compared with the other two baselines.

\section{Conclusions and Future Work}

In this paper, we discuss our proposed SMPACG system for e-commerce platform. The proposed system consists of an automatic product combination selection module and an automatic multi-product advertising copywriting generation module to generate combination with highly related products and corresponding advertising contents automatically. 
The SMPACG has been deployed for serving directly for our e-commerce recommendation system for customers as well as a real-time writing assistant tool for merchants. In future work, we will focus on enhancing the model performance by expanding to multi-modality scenarios.


\clearpage
\bibliography{anthology,custom}
\bibliographystyle{acl_natbib}





\end{document}